\documentclass[10pt,twocolumn,letterpaper]{article}

\usepackage{cvpr}

\usepackage[utf8]{inputenc} 
\usepackage[T1]{fontenc}    
\usepackage{hyperref}       
\usepackage{url}            
\usepackage{booktabs}       
\usepackage{amsfonts}       
\usepackage{nicefrac}       
\usepackage{microtype}      

\usepackage{times}
\usepackage{epsfig}
\usepackage{graphicx}
\usepackage{caption}
\usepackage{subcaption}
\usepackage{algorithm}
\usepackage{algorithmic}
\usepackage{multirow}

\usepackage{verbatim}
\usepackage{amsmath}
\usepackage{amssymb}
\usepackage{overpic}

\newcommand{\R}{\mathbb{R}}
\let\bs=\boldsymbol

\def \saliency {\textup{\saliency}}

\def \path {\mathit{path}}

\def \label {\mathit{label}}

\def \data {\textup{data}}
\def \cons {\textup{cons}}

\numberwithin{theorem}{section}
\numberwithin{lem}{section}

\let\set=\mathcal

\global\long\def\R{\mathbb{R}}
\newcommand{\qixing}[1]{\textcolor{red}{#1}}

\cvprfinalcopy

\def\cvprPaperID{6} 

\title{Joint Learning of Neural Networks via Iterative Reweighted Least Squares}

\author{
Zaiwei Zhang\\
UT Austin
\and
Xiangru Huang\\
UT Austin
\and
Qixing Huang\\
UT Austin
\and
Xiao Zhang\\
Google Inc.
\and
Yuan Li\\
Google Inc.
}

\begin{document}

\maketitle

\begin{abstract}
In this paper, we introduce the problem of jointly learning feed-forward neural networks across a set of relevant but diverse datasets. Compared to learning a separate network from each dataset in isolation, joint learning enables us to extract correlated information across multiple datasets to significantly improve the quality of learned networks. We formulate this problem as joint learning of multiple copies of the same network architecture and enforce the network weights to be shared across these networks. Instead of hand-encoding the shared network layers, we solve an optimization problem to automatically determine how layers should be shared between each pair of datasets. Experimental results show that our approach outperforms baselines without joint learning and those using pretraining-and-fine-tuning. We show the effectiveness of our approach on three tasks: image classification, learning auto-encoders, and image generation.  
\end{abstract}

\section{Introduction}

\begin{figure}[h]
\includegraphics[width=0.48\textwidth]{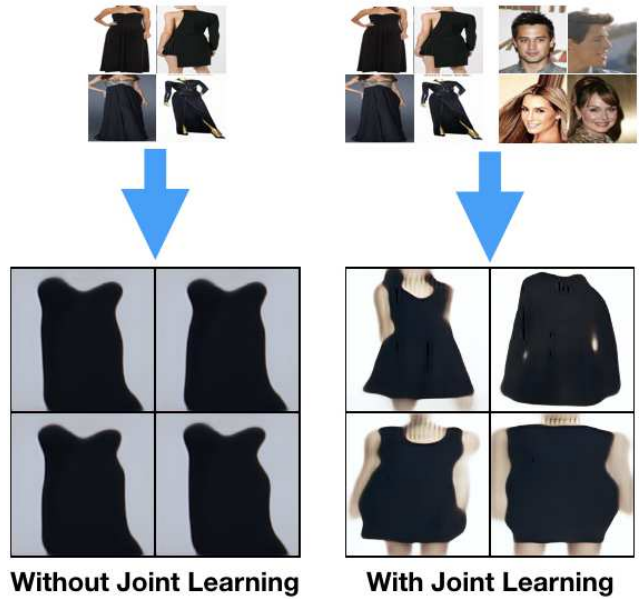}
\caption{Results on black dress image generation with our approach. (Left) Results of an image generator trained from 20K dress images. (Right) Results of an image generator trained from 20K dress images and 200K face images.}
\label{Figure:Teaser}
\end{figure}

Lack of training data remains one of the fundamental challenges in training effective deep neural networks for various visual recognition tasks. One potential solution is to perform transfer learning from other relevant datasets, which essentially amplifies the scale of the training data when training a particular network. A common strategy for transfer learning is to share network weights, i.e., either across the entire network or through a few hand-encoded layers. This strategy has proven to be effective in settings where the input datasets are similar, or where there is prior knowledge informing which layers of features to share. However, in cases where the input datasets exhibit significant variation, this strategy becomes sub-optimal, as it becomes unclear how to determine the shared network layers. 

In this paper, we consider the problem of jointly learning neural networks from a collection of datasets that exhibit significant variations in content and appearance (See Figure~\ref{Figure:Teaser}). We show that despite such significant differences among the input datasets, they still present useful mutual information, which we can use to boost the performance of learning each individual network. We achieve this goal by introducing three novel ways for regularizing the network weights across all datasets. First, instead of hand-encoding the shared network weights, we formulate an optimization problem to assign them automatically. Moreover, rather than enforcing that shared network weights be identical, we use soft constraints to penalize the differences between pairs of shared network layers, allowing us to account for domain shifts between datasets. Finally, we jointly optimize the consistency among network weights using a robust norm, which allows us to extract layer-wise dataset clusters for weight-sharing.   

Specifically, our approach takes a collection of datasets for the same task as input and outputs a learned network for each individual dataset. These networks share the same architecture but have different, yet correlated, layers. We integrate learning shared network layers and learning network weights through a unified optimization problem. The objective function combines a task-specific loss and a consistency term. The consistency term uses a robust norm to automatically determine how each layer should be shared. We also introduce a simple formulation to prioritize that the sharing scheme is consistent across adjacent layers. Our formulation admits effective optimization via iterative reweighted least squares (or IRLS).

We demonstrate the effectiveness of our approach across three diverse tasks: fine-grained image classification, learning auto-encoders, and image generation. These tasks range from predicting a single label (i.e, fine-grained image classification) to dense predictions (i.e., learning auto-encoders and image generation), and from supervised tasks (i.e., fine-grained image classification) to unsupervised tasks (i.e., learning auto-encoders and image generation). Across these tasks, we show that our approach is significantly better than learning each network in isolation, using an L2-norm to share weights, as well as pretraining-and-finetuning. In particular, our approach extracts useful mutual information from datasets that seem to be visually uncorrelated, making our approach suitable across a broad range of settings. 

In summary, we present the following contributions in this paper:
\begin{itemize}
\item We propose to study joint learning of neural networks in the heterogeneous setting, where there exist salient inter-dataset variations. We hope our method inspires further research along this direction. 
\item We propose a robust norm to automatically determine how to share layers between each pair of datasets.
\item We demonstrate the effectiveness of our approach on three tasks, including image classification, learning auto-encoders, and image generation. 
\end{itemize}
\section{Related Works}

\noindent\textbf{Domain adaptation.} Our problem falls in the general category of domain adaptation. It is beyond the scope of this paper to provide a comprehensive review of the literature. We refer to ~\cite{Busto_2017_ICCV,Gebru_2017_ICCV,Gholami_2017_ICCV,Carlucci_2017_ICCV,Tzeng_2017_CVPR} for recent advances and to~\cite{DBLP:journals/corr/Csurka17} for a recent survey on this topic. For discriminative tasks (e.g., image classification and image segmentation), visual domain adaptation techniques fall into supervised domain adaptation techniques~\cite{Chopra:2005:LSM,DBLP:journals/corr/RozantsevSF16} and unsupervised domain adaptation techniques~\cite{Tzeng_2017_CVPR,DBLP:journals/corr/KimCKLK17,DBLP:journals/corr/RussoCTC17}. Our approach falls into the supervised category. A common approach for supervised techniques is to share weights across networks. This strategy works for the case where the domain shifts are relatively small, but tends to break when the domain shifts are large. A potential solution is to share weights across a subset of predefined layers (e.g., \cite{Su_2015_ICCV}). However, this method requires prior knowledge of the mutual information across datasets, and we found that such information is not obvious, particularly between visually dissimilar datasets. In addition, it becomes extremely difficult to hand-encode weights across multiple networks. Instead, we solve a joint optimization problem to determine the matched layers. Matching network weights under the $L^2$ norm has been considered in a recent work~\cite{DBLP:journals/corr/RozantsevSF16}. The difference in this work is that we propose to use robust norms to automatically extract matched network weights between pairs of datasets, and we do so across multiple datasets in a consistent manner. 

State-of-the-art unsupervised domain adaptation techniques build maps across the domains~\cite{DBLP:journals/corr/KimCKLK17,DBLP:journals/corr/RussoCTC17,DBLP:journals/corr/ZhuPIE17}. In particular, the inter-domain maps are enforced to be consistent. These maps are usually trained using generative adversarial networks~\cite{NIPS2014_5423} and training procedures described in follow-up works~\cite{DBLP:journals/corr/SalimansGZCRC16,DBLP:journals/corr/ArjovskyCB17}. Although we do not consider unsupervised domain adaptation in this paper, our approach can be potentially used for joint training of the discriminators used in each domain. 

For synthesis tasks such as image generation, Liu et al.~\cite{NIPS2016_6544} proposed a method for training a pair of generative adversarial networks. Their strategy is similar to supervised domain adaptation for image classification/segmentation, i.e., by sharing hand-encoded layers. In contrast, our approach automatically learns matched layers, and we do so across multiple domains. 

\noindent\textbf{Joint object matching.} Joint linking of corresponding layers across multiple networks is related to a recent line of work on joint optimization of object maps among image and shape collections~\cite{DBLP:journals/tog/HuangZGHBG12,DBLP:journals/cgf/HuangG13}. Similar to our setting, the central theme of these methods is to enforce the consistency of maps along cycles, so that a noisy map between two different objects can be computed by composing maps along a path of similar objects. In this paper, we apply this methodology to link layers of neural networks. Another related line of work~\cite{zhou2016learning,DBLP:journals/corr/KimCKLK17} focuses on utilizing the cycle-consistency constraint for regularization when training neural networks. Our problem differs from these works in that the neural networks optimized in our setting are associated with each dataset, and we establish consistent correspondences between network weights. In contrast, in these works the neural networks are defined between pairs of datasets. 


\section{Problem Setup}
\label{Section:Problem:Setup}

In this Section, we formally define the joint neural network learning problem we consider in this paper. Suppose we are given $n$ datasets for a specific task (e.g., image classification) and a neural network $G_{\theta}$ designed for it. Without losing generality, we assume $G_{\theta}$ is a feed-forward neural network that consists of $L$ layers $\theta^{1},\cdots, \theta^{L}$, and each layer $l$ has $n_{l}$ parameters. However, our approach can be easily adapted for more sophisticated networks that connect layers using a graph. Our goal is to learn $n$ network parameters $\theta_1,\cdots, \theta_{n}$, one for each dataset. Instead of learning them separately, we propose to learn them jointly. Unlike previous approaches that hand-encode the layers with shared parameters, our approach solves an optimization problem to simultaneously optimize the network parameters and determine the shared network layers. 

Without losing generality, we denote the loss function from dataset $\set{I}_i$ as $f_i(\theta_i)$. In Section~\ref{Section:Approach}, we will use $f_i(\theta_i)$ to present both the formulation and the optimization procedure. In the following, we present the explicit expressions of $f_i(\theta_i)$ for the three tasks considered in this paper, namely, fine-grained image classification, learning auto-encoders, and learning generative models.

\subsection*{Task-Specific Loss Terms}

\noindent\textbf{Classification Loss.} The first task we consider is image classification. In this setting, each dataset is given by a set of labeled images $\set{I}_i = \{(I, y_{I})\}$, where $y_{I}$ is the labeled associated with $I$. The corresponding data dependent loss function is then given by
\begin{equation}
\resizebox{0.25\textwidth}{!}{$
f_i(\theta_i) = \frac{1}{|\set{I}_i|}\sum\limits_{(I, y_I)\in \set{I}_i}\ l(G_{\theta_i}(I), y_I),
$} \nonumber
\end{equation}
where we set $l(\cdot, \cdot)$ as the cross entropy loss between predicted labels and ground-truth labels. 


\noindent\textbf{Auto-Encoder Loss.} The second task is training an auto-encoder for a collection of images. We will evaluate auto-encoders indirectly, e.g., through their reconstruction loss on testing images and in the application of image completion. In this setting, each dataset $\set{I}_i = \{I\}$ is given by a collection of unlabeled images. Following~\cite{DBLP:journals/corr/KingmaW13}, we directly use the regression loss to define the auto-encoder loss:
\begin{equation}
\resizebox{0.25\textwidth}{!}{$
f_i(\theta_i) = \frac{1}{|\set{I}_i|}\sum\limits_{I \in \set{I}_i}\ \|I - G_{\theta_i}(I)\|_{\set{F}}^2.
$} \nonumber
\end{equation}

\noindent\textbf{Generative Adversarial Loss.} The third task is training a generative adversarial network for a collection of real images. We will adopt the BEGAN~\cite{berthelot2017began} architecture, where we share the auto-encoders across different domains, and also adopt the DCGAN~\cite{radford2015unsupervised} architecture, where we share all the network parameters across different domains. Following~\cite{goodfellow2014generative}, we use the adversarial loss to optimize the generative models:
\begin{equation}
\resizebox{0.4\textwidth}{!}{$
f_i(\theta_i) = \frac{1}{|\set{I}_i|}\sum\limits_{I \in \set{I}_i}\ (L(D_{\theta_i}(I)) - L(D_{\theta_i}(G_{\theta_i}(z|_{z \in \mathcal{N}})))),
$} \nonumber
\end{equation}
where $L(\cdot)$ depends on the neural networks used. and are discussed in Section~\ref{Section:Results}.

\section{Approach}
\label{Section:Approach}

We proceed to present the proposed approach for joint learning of neural networks. In Section~\ref{Subsection:Formulation}, we describe the proposed formulation. Then in Section~\ref{Subsection:Optimization}, we show how to effectively solve the induced optimization problem. 

\subsection{Formulation}
\label{Subsection:Formulation}

The proposed formulation combines a data term $f_{\data}$ and a consistency term $f_{\cons}$. The data term $f_{\data}$ simply adds the loss from each dataset together:
\begin{equation}
f_{\data} = \sum\limits_{i=1}^{n} f_i(\theta_i).
\label{Eq:Data:Term}
\end{equation}

The regularization term $f_{\cons}$ forces the network parameters to match. In the presence of diverse datasets, the desired layer-wise network parameters $\theta_i^{l}, 1\leq i \leq n, 1\leq l \leq L$ shall possess the following  properties :
\begin{itemize}
\item For each layer $l$, the input datasets form clusters so that for two datasets $I_i$ and $I_j$ that belong to the same cluster, $\theta_i^{l}$ and $\theta_j^{l}$ shall be close to each other. The motivation comes from the success of sharing bottom or top layers between a pair of networks for various domain adaption tasks (e.g.,\cite{NIPS2016_6544,Su_2015_ICCV}). In this paper, we generalize it to multiple networks. However, we do not assume the underlying clusters are given.
\item The cluster structures are consistent between consecutive layers. This property is also motivated from the common practice of sharing a block of consecutive layers between a pair of networks (e.g.,\cite{NIPS2016_6544}). Again, we do not assume these blocks are given.
\end{itemize}

Our formulation of $f_{\cons}$ is motivated from \cite{DBLP:journals/corr/abs-1709-01870}, which perform data clustering using robust fusion penalties. Specifically, given a set of points $\bs{p}_1,\cdots, \bs{p}_n \in \R^{d}$, these approaches solve the following problem to find perturbed points $\bs{x}_1,\cdots, \bs{x}_n \in \R^d$ (c.f.~\cite{DBLP:journals/corr/abs-1709-01870}):
\begin{equation}
\min\limits_{\bs{x}_1,\cdots, \bs{x}_n}\sum\limits_{i=1}^{n}\|\bs{p}_i-\bs{x}_i\|^2 + \sum\limits_{1\leq i < j \leq n}\rho(\|\bs{x}_i-\bs{x}_j\|),
\end{equation}
where $\rho(\cdot)$ is a robust norm. In~\cite{DBLP:journals/corr/abs-1709-01870}, the authors have shown that the perturbed locations of the same cluster tend to be identical. 

We adapt this formulation to prioritize that the network parameters at each layer form clusters. To ensure that the cluster structures are consistent between adjacent layers, our key idea is to apply the formulation on concatenated layer-wise parameters $(\theta_i^l, \theta_i^{l+1})$. In other words, we would like to cluster $(\theta_i^l, \theta_i^{l+1})$ together, which implicitly forces the cluster structures to be consistent across adjacent layers:
\begin{equation}
f_{\cons} = \sum\limits_{1\leq i < j \leq n} \sum\limits_{l=1}^{L-1} \rho(\|(\theta_i^{l}, \theta_i^{l+1})-(\theta_j^{l}, \theta_j^{l+1})\|, \sigma^{l}).
\label{Eq:Regularization:Term}
\end{equation}
Here we choose a variant of the Huber loss to define the robust norm
$$
 \rho(x, \sigma) = \frac{\sigma^2 x^2}{\sigma^2 + x^2},
$$
where $\sigma$ is a hyper-parameter that determines the transition
from $x^2$ to $\sigma^2$ when increasing $x$. We will discuss how to set the hyper-parameters $\sigma^{l}, 1\leq l \leq L$ in Section~\ref{Subsection:Optimization}. 

Combining (\ref{Eq:Data:Term}) and (\ref{Eq:Regularization:Term}), we arrive at the following optimization formulation for joint learning of neural networks:
\begin{equation}
\resizebox{0.45\textwidth}{!}{$
\min\limits_{\{\theta_i\}}\ \sum\limits_{i=1}^{n} f_i (\theta_i) + \lambda \sum\limits_{1\leq i,j\leq n}\sum\limits_{l=0}^{L-1}\rho(\|(\theta_i^l, \theta_i^{l+1})-(\theta_j, \theta_j^{l+1})\|, \sigma^l)$}
\label{Eq:Objective:Function}
\end{equation}
In this paper, we choose $\lambda = 10$ for all the experiments. We also find that due to the usage of the robust norm, the performance of the resulting networks is insensitive to the values of $\lambda$.

\subsection{Optimization}
\label{Subsection:Optimization}

\begin{figure}[t]
\centering
\includegraphics[width=0.45\textwidth]{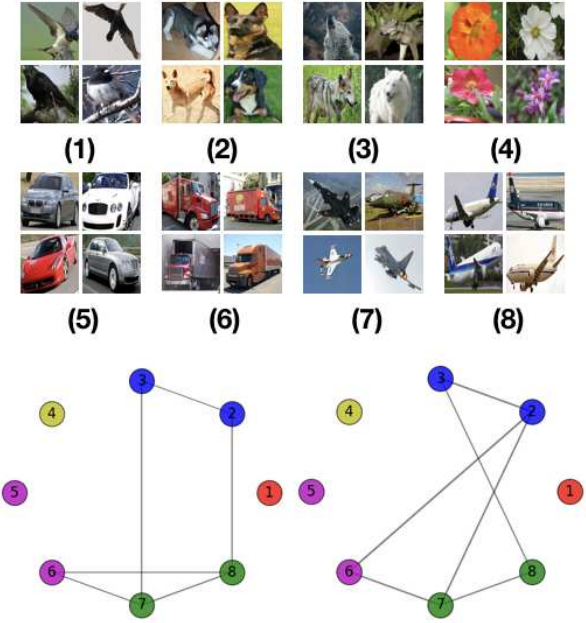}
\caption{``Social network" formed by 8 datasets in two representative layers. For the first few layers, we found that all datasets share similar parameters. For middle layers of encoder and decoder networks, we construct a graph by adding an edge between $i$ and $j$ if and only if both of them are one of the 3 most influential neighbors of each other. A community of mutual influence can be observed.
(Left) Middle layers of encoder network, (Right) Middle layers of decoder network. Between similar datasets, such as Dog(2) and Wolf(3) or Fighter Aircraft(7) and Plane(8), more weights are shared compared to other pairs. However, some connections are not intuitive, such as Truck(6) and Fighter Aircraft(7), the system finds it useful to decrease the overall objective.}
\label{Fig:LayerWise:Social:Network}
\end{figure}
Motivated from the success of applying iterative reweighted least squares (or IRLS) for minimizing robust norms (e.g.,~\cite{Daubechies_iterativelyreweighted,DBLP:conf/iccv/ChatterjeeG13}), we adapt IRLS to solve (\ref{Eq:Objective:Function}). In the following, we first describe how to initialize the network parameters, and how to set the hyper-parameters $\sigma^l, 1\leq l \leq L$. We then present the IRLS procedure for solving (\ref{Eq:Objective:Function}).

\subsubsection{Initializing Network Parameters and Determining Hyper-Parameters}

We follow the standard practice in IRLS, which initializes the network parameters by replacing the robust norm with the L2-norm:
\begin{equation}
\resizebox{0.45\textwidth}{!}{$
\min_{\{\theta_i\}}\ \sum\limits_{i=1}^{n} f_i(\theta_i) + \lambda \sum\limits_{1\leq i<j\leq n}\sum\limits_{l=0}^{L-1}\|(\theta_i^l, \theta_i^{l+1})-(\theta_j^l, \theta_j^{l+1})\|^2 $}
\label{Eq:1}
\end{equation}
(\ref{Eq:1}) can be reformulated as a special case of minimizing a general objective function:
\begin{equation}
\resizebox{0.4\textwidth}{!}{$
\underset{\theta_i, 1\leq i \leq n}{\min}\ \sum\limits_{i=1}^{n} f_i(\theta_i) +  \sum\limits_{1\leq i<j \leq n}\sum\limits_{l=1}^{L} c_{ijl}\|\theta_{i}^{l}-\theta_{j}^{l}\|^2 $}
\label{Eq:Network:Opt}
\end{equation}
where $c_{ijl} > 0$ are constants. As we will see later, (\ref{Eq:Network:Opt}) can also be used for solving intermediate steps of IRLS. To avoid breaking the flow of the paper, we defer the technical details for solving (\ref{Eq:Network:Opt}) to Section~\ref{Subsubsection:Joint:Network:Optimization}.

Let ${{\theta}_i^l}^{(0)}, 1\leq i \leq n, 1\leq l \leq L$ be the resulting network parameters from (\ref{Eq:Network:Opt}). We set the layer-wise hyper-parameter as
\begin{equation}
\resizebox{0.35\textwidth}{!}{$
\sigma^{l} = \underset{1\leq i \leq n}{\textup{mean}}\ \min\limits_{j \neq i}\|({\theta_i^l}^{(0)}, {\theta_i^{l+1}}^{(0)})-({\theta_j^{l}}^{(0)}, {\theta_j^{l+1}}^{(0)})\|, \nonumber
$}
\end{equation}
which works well in all of our experiments. 

\subsubsection{Reweighted Least Squares Regularization}

We proceed to apply IRLS to minimize the objective function in (\ref{Eq:Objective:Function}). Each iteration of IRLS consists of a weighting step and an optimization step. In this paper, we consider splitting the robust norm as $\rho(x, \sigma) = \frac{\sigma^2}{\sigma^2 + x^2}\cdot x^2$, leading to the following weighting-optimization procedure:

\noindent\textbf{Weighting.} Denote the network parameters at iteration $k-1$ as ${\theta_i^{l}}^{(k-1)}$. We introduce a weight $w_{ijl}^{(k)}$ for term $\rho(\|(\theta_i^l,\theta_i^{l+1})-(\theta_j^l,\theta_j^{l+1})\|, \sigma^l)$ at iteration $k$ as
\begin{equation}
\resizebox{0.3\textwidth}{!}{$
w_{ijl}^{(k)} = \frac{{\sigma^l}^2}{{\sigma^l}^2 + \|(\theta_i^l,\theta_i^{l+1})-(\theta_j^l,\theta_j^{l+1})\|^2}.
$}
\end{equation}
Intuitively, the dataset pairs that are further away from each other at layer $l$ will be associated with small weights. 

\noindent\textbf{Optimization.} After determining the weights, we modify (\ref{Eq:Objective:Function}) and solve the following optimization problem:
\begin{equation}
\resizebox{0.45\textwidth}{!}{$
\{\theta_i^{(k+1)}\} = \min\limits_{\{\theta_i\}}\ \sum\limits_{i=1}^{n} f_i (\theta_i) + \lambda \sum\limits_{1\leq i,j\leq n}\sum\limits_{l=0}^{L-1}w_{ijl}^{(k)}\|(\theta_i^l, \theta_i^{l+1})-(\theta_j^{l}, \theta_j^{l+1})\|^2$}
\label{Eq:Objective:Function2}
\end{equation}
(\ref{Eq:Objective:Function2}) is again a special case of (\ref{Eq:Network:Opt}), and we will discuss the optimization procedure in Section~\ref{Subsubsection:Joint:Network:Optimization}. 

We can understand the behavior of IRLS as follows. For layers with similar parameters, the corresponding weights are close to $1$, while for layers with dissimilar parameters, the corresponding weights are close to $0$. With such weights, solving (\ref{Eq:Objective:Function2}) would push the layers with similar weights to be even closer to each other. In the mean-time, layers with dissimilar weights are likely to be pulled away from each other due to the data terms. In the end, layers tend to form clusters.

Across all of our experiments, we found that IRLS converges favorably fast. In our implementation, we monitor 
\begin{equation}
\resizebox{0.3\textwidth}{!}{$
\delta^{(k)}:= \max\limits_{1\leq i<j \leq n, 1\leq l \leq L}\ |w_{ijl}^{(k)}-w_{ijl}^{(k-1)}|,$} \nonumber
\end{equation}
and terminate the IRLS procedure when $\delta^{(k)} \leq 10^-2$. In our experiments, we found 4-8 iterations are sufficient for convergence. 

Figure~\ref{Fig:LayerWise:Social:Network} illustrates the links that connect layers with similar optimized parameters over eight datasets for the task of learning auto-encoders. We can see that the links reveal meaningful shared information across the datasets.  

\begin{figure}
\centering

\def\arraystretch{0}
\begin{tabular}{@{}*5{c}@{}}
\hspace{-0.25cm}
\begin{overpic}[width=0.24\textwidth]{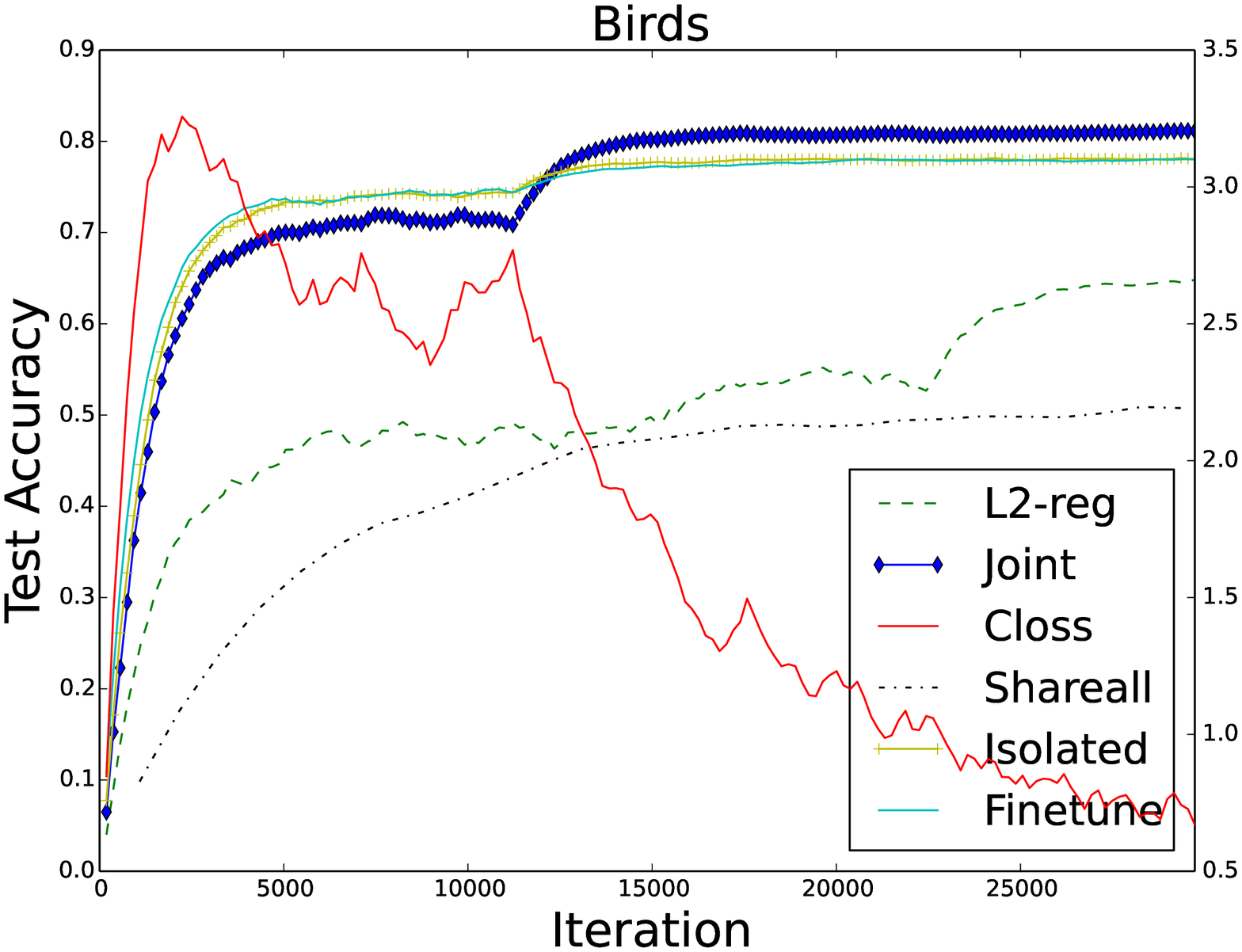}
\end{overpic}
\begin{overpic}[width=0.24\textwidth]{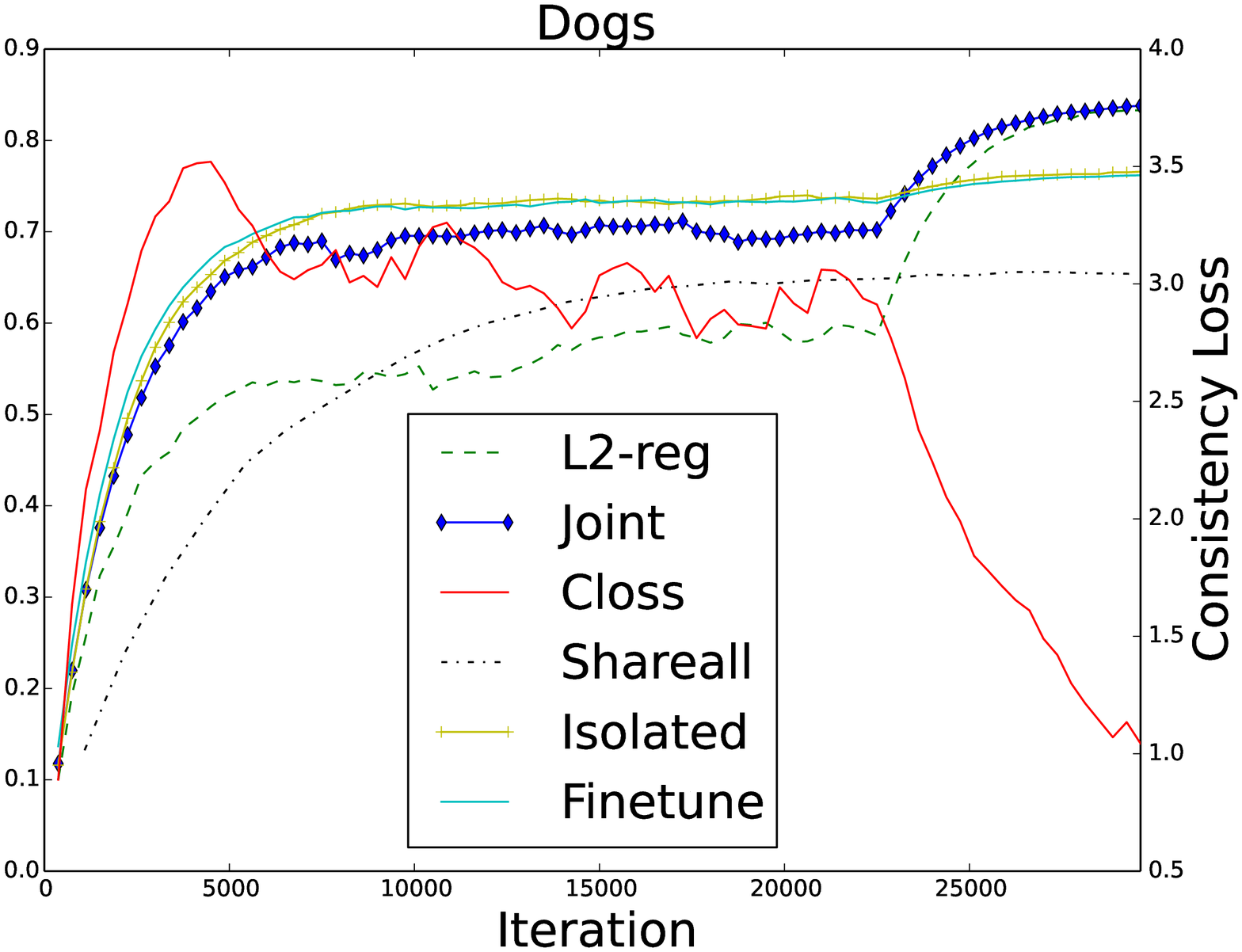}
\end{overpic}
\\

\end{tabular}


\vspace{0.03in}
\caption{ For the Image Classification task, moving average of test accuracy vs number of training iterations is plotted for two datasets: 
Birds, Dogs (from left to right). Compared baselines are Isolated-Training (Isolated), L2-regularization (L2-reg), Shareall, Pretrain-Finetune (Finetune). Our method is indicated as Joint. Consistency term loss (Closs) vs number of training iterations for Joint method is also plotted.}
\label{Figure:Image:Classification}
\end{figure}

\subsubsection{Joint Network Optimization}
\label{Subsubsection:Joint:Network:Optimization}

In this section, we describe the technical details for solving (\ref{Eq:Network:Opt}), which has been used in the variable initialization stage as well as the reweighted least squares stage. When $n$ is large, it is hard to solve (\ref{Eq:Network:Opt}) directly, since it involves one copy of the network for each dataset. Motivated from the success of block-coordinate descent techniques for solving large-scale optimization problems (c.f.~\cite{Boyd:2011:DOS}), we propose to solve (\ref{Eq:Network:Opt}) by optimizing one network at a time while fixing the other networks. Specifically, suppose we pick the $i$-th network at the current iteration, and with ${\overline{\theta}_j^{l}}, j\neq i, 1\leq l \leq L$ we denote the current parameters of other networks. Then it is clear that (\ref{Eq:Network:Opt}) reduces to
\begin{equation}
\min\limits_{\theta}\ f_i(\theta) + \lambda \sum\limits_{l=1}^{L} \sum\limits_{j \neq i} c_{ijl}\|\theta^l - \sum\limits_{j \neq i}c_{ijl}\overline{\theta}_j^{l}/\sum\limits_{j \neq i}c_{ijl}\|^2  
\label{Eq:Block:Coordinate:Descent}
\end{equation}
Since $\sum\limits_{j \neq i}c_{ijl}\overline{\theta}_j^{l}/\sum\limits_{j \neq i}c_{ijl}$ is a constant vector when optimizing $\theta_i$, (\ref{Eq:Block:Coordinate:Descent}) can be considered as a standard network training problem, and we apply stochastic coordinate descent for optimization. At each iteration, we train (\ref{Eq:Block:Coordinate:Descent}) with one epoch before moving to the next iteration.

\begin{figure}[t]
\includegraphics[width = 0.48\textwidth]{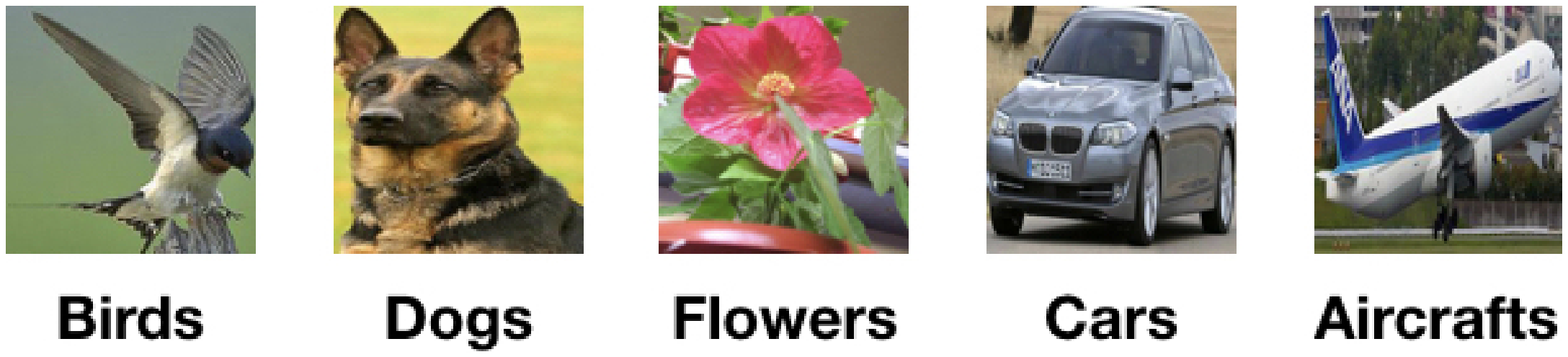}
\begin{tabular}{p {1.4cm}| p {0.65cm} | p {0.65cm} | p {0.8cm} | p {0.63cm} | p {0.87cm} | p {0.5cm} } 
 \hline
 \footnotesize{Method} & \footnotesize{Birds} & \footnotesize{Dogs} & \footnotesize{Flowers} & \footnotesize{Cars} & \footnotesize{Aircrafts} & \footnotesize{Avg} \\
 \hline\hline
  \footnotesize{Joint(ours)} & \footnotesize{\textbf{81.08}} & \footnotesize{\textbf{84.31}} & \footnotesize{\textbf{95.97}} & \footnotesize{91.43} & \footnotesize{84.28} & \footnotesize{\textbf{87.41}} \\ 
 \hline
 \footnotesize{Isolated} & \footnotesize{78.06} & \footnotesize{76.70} & \footnotesize{95.95} & \footnotesize{\textbf{92.33}} & \footnotesize{84.62} & \footnotesize{85.53}\\ 
 \hline
 \footnotesize{L2Reg} & \footnotesize{64.92} & \footnotesize{83.40} & \footnotesize{52.38} & \footnotesize{66.94} & \footnotesize{27.19} & \footnotesize{58.97}\\
 \hline
  \footnotesize{L2Reg(25\%)} & \footnotesize{62.52} & \footnotesize{64.83} & \footnotesize{67.84} & \footnotesize{82.54} & \footnotesize{68.30} & \footnotesize{69.21}\\
 \hline
  \footnotesize{L2Reg(50\%)} & \footnotesize{67.97} & \footnotesize{67.84} & \footnotesize{69.38} & \footnotesize{79.72} & \footnotesize{65.87} & \footnotesize{70.16}\\
 \hline
 \footnotesize{Shareall} & \footnotesize{58.03} & \footnotesize{66.07} & \footnotesize{80.39} & \footnotesize{68.99} & \footnotesize{51.89} & \footnotesize{65.07}\\
 \hline
 \footnotesize{Finetune} & \footnotesize{78.02} & \footnotesize{76.31} & \footnotesize{95.31} & \footnotesize{91.83} & \footnotesize{\textbf{84.86}} & \footnotesize{85.27}\\
 \hline
\end{tabular}
\caption{Classificaiton accuracy on testing datasets. Compared baselines are: Isolated-Training (Isolated), L2-regularization (L2Reg), L2Reg with 25\% weight shared, L2Reg with 50\% weight shared, Shareall, and Pretrain-Finetune(Finetune). Our approach achieves the best overall performance across all the datasets. }
\label{table:Classification}
\end{figure}
\section{Experimental Evaluation}
\label{Section:Results}

In this section, we present the experimental evaluations of the proposed approach. We first describe the experimental setup. We then evaluate the benefits of the proposed approach on each specific task. Please refer to the supplemental material for additional results.

\subsection{Experimental Setup}

\noindent\textbf{Datasets.} We summarize the datasets we used for each specific task below. Please refer to the supplemental material for a detailed specification:
\begin{itemize}
\item \textbf{Fine-grained image classification.} For this task, we consider five fine-grained classification datasets: Cars Dataset ~\cite{krause20133d}, FGVC-Aircraft Benchmark ~\cite{maji13fine-grained}, Stanford Dogs Dataset ~\cite{KhoslaYaoJayadevaprakashFeiFei_FGVC2011}, Caltech-UCSD Birds-200-2011 ~\cite{WahCUB_200_2011}, 102 Category Flower Dataset ~\cite{Nilsback08} for evaluating our joint learning approach. Note that the appearance of these images are considered visually dissimilar. We pick fine-grained classification tasks to demonstrate that across different domains, our approach can still extract meaningful mutual information, leading to performance gains. For this task, we employ InceptionV3~\cite{szegedy2016rethinking} network architecture, with pretrained weights from ImageNet for initialization. In order to mitigate over-fitting, we carry out the Inception-style data augmentation methods: scale and aspect ratio variation, and image distortion. 

\item \textbf{Auto-encoder learning.} The second task considers image completion from multiple domains of images with auto-encoders. For this task, we employ a 9-layer auto-encoder architecture (details are deferred to the supplemental material). We picked images from 8 classes (image samples are shown in Figure \ref{Fig:LayerWise:Social:Network}), which are Bird, Dog, Wolf, Flower, Car, Truck, Plane, and Fighter Aircrafts, using 500 images for training and 100 for testing. All images are retrieved from ImageNet, and sampled from the five fine-grained classification datasets. For each image, we randomly cropped out a 22x22 square for the image completion task.

\item \textbf{Image generation.} In the third task we consider image generation using a collection of images from different domains. For this task, we employ the experimental setting (e.g., network architecture) from both ~\cite{radford2015unsupervised} and ~\cite{berthelot2017began}. We have conducted three experiments which involve six datasets. The first dataset is comprised of video snapshots taken from ~\cite{ActionsAsSpaceTimeShapes_iccv05}, and contains 3K images of humans walking down a street, while the second dataset contains 10K synthetic images of humans walking, which we generated. The third dataset is CelebA~\cite{liu2015faceattributes}, which contains around 200K images, and the fourth dataset is a collection of 20K black dress images crawled from the web. We additionally use the Flower dataset~\cite{Nilsback08} and the Bird dataset~\cite{WahCUB_200_2011}.
\end{itemize}

\noindent\textbf{Baseline Approaches.} We consider the following four baselines to assess the proposed approach.
\begin{itemize}
\item \textbf{Baseline I: Isolated-Training.} The first baseline simply trains the network from each dataset independently. To make a fair comparison between the proposed approach and state-of-the-art approaches, we report the performance of state-of-the-art methods that use AlexNet with weights pre-trained on ImageNet. 

\noindent\textbf{Baseline II: L2-regularization} The second baseline replaces the robust-norm by the L2-norm for regularizing differences in layer-wise parameters. This baseline assesses the benefit of using the robust norm for regularization.

\noindent\textbf{Baseline III: Shareall.} The third baseline simply shares the same network parameters across all the datasets. This baseline is introduced to assess the importance of sharing weights in an adaptive manner. 

\noindent\textbf{Baseline IV: Pretrain-Finetune.} The fourth baseline is generalized from the popular pretraining-fine-tuning paradigm. In this case, we first train a joint model from all the datasets by sharing weights. We then fine-tune the joint model on each dataset independently. This baseline is introduced to assess the advantage of solving an optimization problem for joint training. 
\end{itemize}

\begin{figure}[t]
\centering
\includegraphics[width=0.45\textwidth]{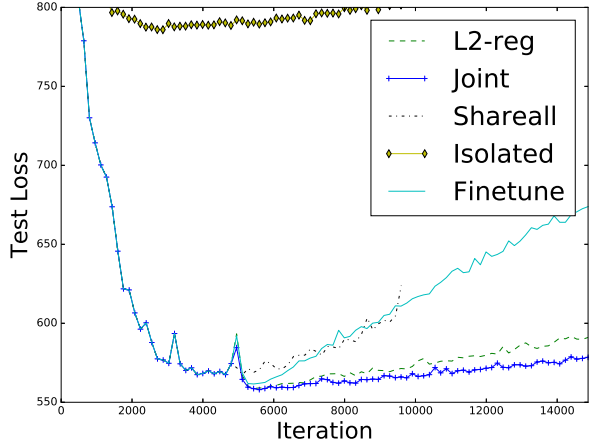}
\caption{For the Image Completion task, average L2 image reconstruction loss for eight datasets vs number of training iterations is plotted. Compared baselines follow the same standard in Figure \ref{Figure:Image:Classification}. We can see our Joint approach leads to the lowest testing accuracy.}
\label{Figure:Data:AutoEncoder_Evaluation}
\end{figure}

\begin{figure*}[t]
\begin{tabular}{@{}c@{}}
\hspace{-0.5cm}
\begin{overpic}[width=1.0\textwidth]{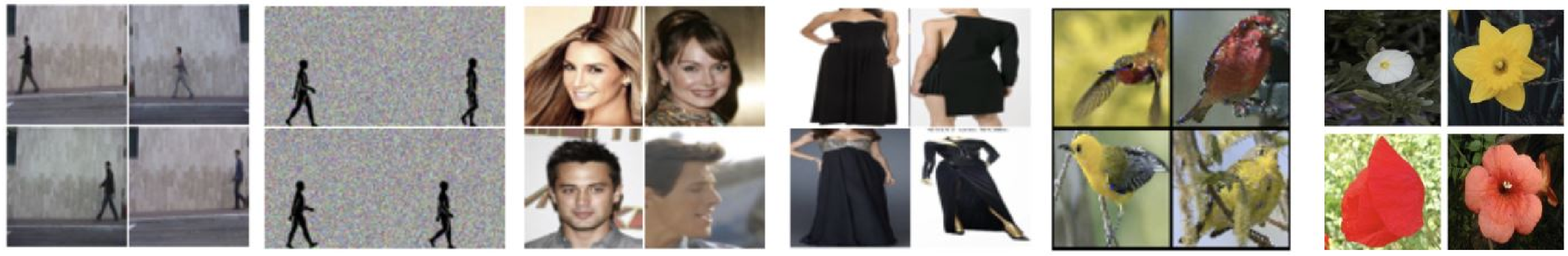}
\end{overpic}\\
\vspace{0.1in}
\hspace{-1.0cm} Random dataset samples \\
\hspace{-0.5cm}
\begin{overpic}[width=1.0\textwidth]{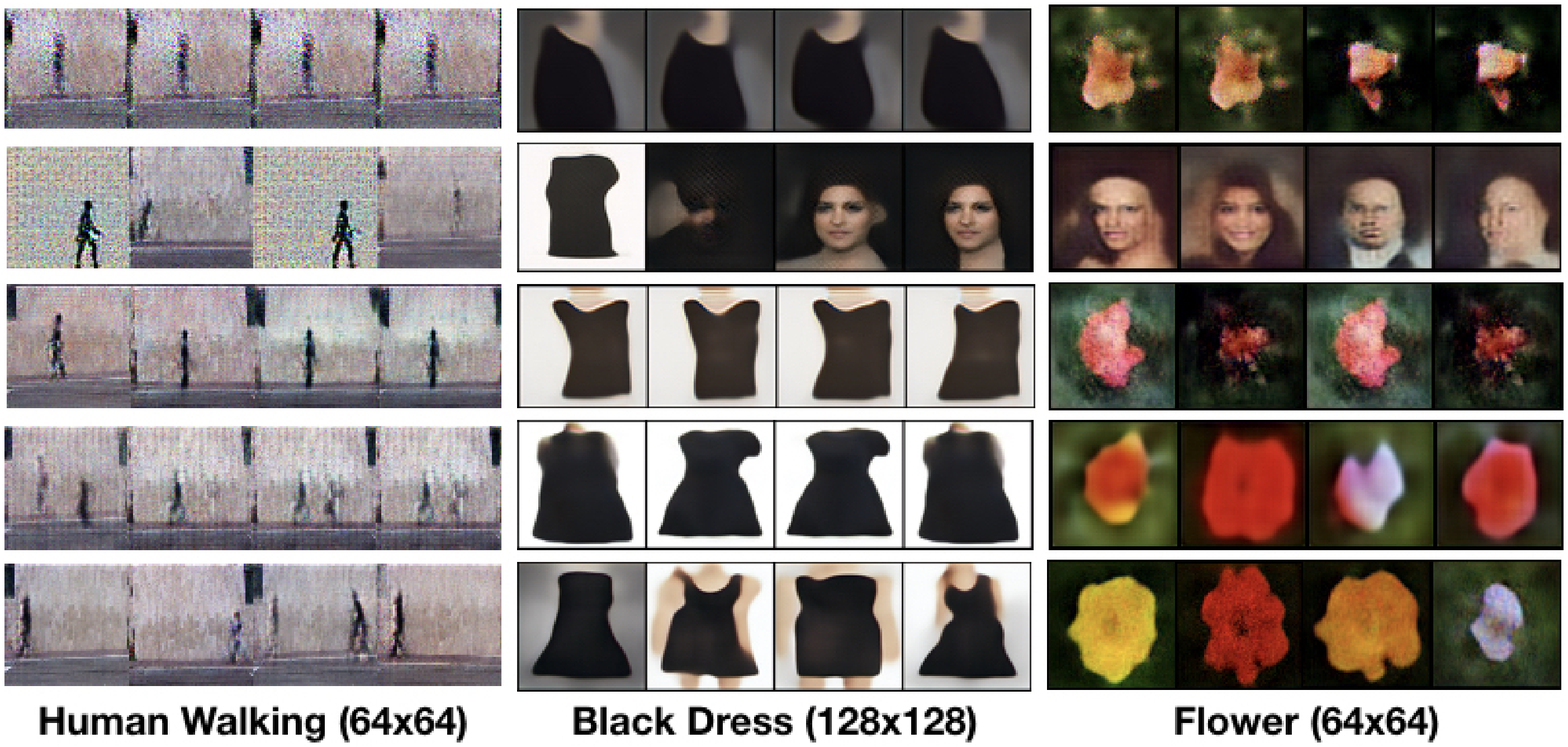}
\end{overpic} \\
\vspace{-0.15in}
\hspace{-0.75cm}
\end{tabular}
\caption{\textbf{Image generation results.} This figure shows the image generation results on three groups of datasets: Human Walking (Real and Synthetic), Black Dress+Faces, and Flower+Bird. (Top Block) 4 Samples from each dataset in each group. (Bottom Block) Image generation results. From top row to bottom row: we show Isolated, Shareall, Pretrain-finetune, L2-regularization and our approach.}
\label{Figure:Image:Generation}
\end{figure*}

\subsection{Task I: Image Classification.}
\label{Subsection:Image:Classification}

Table~\ref{table:Classification} illustrates the testing accuracy of our approach and baseline approaches. Our approach outperforms most baseline approaches across all the input datasets. The improvement over Isolated-Training is significant in most cases, especially on Birds and Dogs dataset. Figure \ref{Figure:Image:Classification} shows the performance gain during training for those two datasets, and only when consistency loss drops, there are significant boosts in testing accuracy for Joint method. This demonstrates the importance of sharing weights. L2-Regularization improves over Isolated-Training in some cases. However, falsely linking datasets together causes unstable behaviors, such as very low testing accuracy on Aircrafts. Using the robust norm not only lifts the testing accuracy but also insures at least similar performance compared to Isolated-Training. This shows the advantage of using a robust norm for regularization in the presence of diverse datasets. A surprising result is that Shareall leads to the lowest accuracy in some cases. This can be understood from the fact that the five datasets are diverse, and it is important to allow some difference across the networks (e.g., using L2-Regularization). Finally, Pretrain-Finetune improves over Shareall but only matches the performance of Isolated-Training due to the quality of the pretrained weights from Shareall. To test that our automatic weight sharing scheme is better than sharing hand-crafted features, we also propose two more baseline comparisons: L2-Regularization with 25\% and 50\% weight shared in the first few layers. Our automatic weight sharing scheme outperforms those baseline methods by more than 15\% in average. Code is publicly available at \href{https://github.com/zaiweizhang/Joint-Learning-of-NN}{here}.

\subsection{Task II: Learning Auto-Encoders}
\label{Subsection:Auto:Encoders}

Figure~\ref{Figure:Data:AutoEncoder_Evaluation} illustrates the average reconstruction loss of our approach and baseline approaches on the testing data (which is left out during training). Compared to Isolated-Training and Pretrain-Finetune, our joint learning method leads to the smallest reconstruction error on the testing data. This again shows the importance of sharing weights in a soft manner and using a robust norm to filter out irrelevant layers.  In particular, compared to Shareall, the performance gain is significant. Moreover, Pretrain-Finetune results in overfitting. The improvements over the other two baselines are also noticeable. For example, the generalization behavior of Isolated remains poor. Figure~\ref{Fig:LayerWise:Social:Network} illustrates links that connect layers with similar weights. We can see that our approach successfully recover the intrinsic similarities across the datasets, e.g., between the two animal datasets and between the two airplane datasets. 


\subsection{Task III: Image Generation}

We conducted the following three experiments on the image generation task:
\begin{itemize}
\item Use DCGAN~\cite{radford2015unsupervised} to generate images of humans walking down a street by joint learning with 3K real and 10K synthetic images.

\item Use BEGAN~\cite{berthelot2017began} to generate images of black dresses by joint learning with 20K dress and 200K face images.

\item Use BEGAN~\cite{berthelot2017began} to generate images of flowers by joint learning with 2K flower, 6K bird, and 200K face images.
\end{itemize}

Figure~\ref{Figure:Image:Generation} shows the qualitative results of the image generation task. We can see that our approach leads to the best general results across the input datasets. As for baseline approaches, Isolated training easily overfits the training data. As the datasets are quite diverse, Shareall, which uses one network to generate images, tends to generate images that interpolate across different categories. The resulting images are thus unrealistic. Starting from Shareall and then finetuning on each dataset certainly improves the visual quality. However, the results are still not as competent as our approach. L2-regularization, which uses a soft weight sharing scheme, avoids the issue of generating mixed-class images experienced by Shareall. However, the issue of L2-regularization is that it evenly distributes the error, resulting in overly smooth images.

\section{Conclusions and Future Work}

In this paper, we have introduced a method for joint learning of neural networks among a collection of relevant and diverse datasets for the task of transfer learning. The key idea behind our approach is to use a robust norm to automatically identify which layers should be matched between each pair of datasets. Our formulation also enforces consistency of matches between adjacent layers. The formulation can be easily optimized via iterative reweighted least squares. Experimental results show the advantage of our approach against four baseline approaches, namely, (1) training from each dataset in isolation, (2) sharing the same weights across all the networks, (3) finetuning from shareall, and (4) using the L2-regularization. The improvements are consistent among the three tasks introduced in this paper, namely, fine-grained image classification, learning auto-encoders, and image generation. 

There are ample opportunities for future research. We would like to apply our method for other tasks such as depth prediction and semantic segmentation. Moreover, we would like to apply our approach to other types of neural networks such as recurrent neural networks and densely connected networks. Moreover, we have used the same network architecture across all datasets. It would be interesting to extend the approach to jointly learn different but correlated networks, e.g., networks for image segmentation and networks for image classification may share similar convolutional layers. Finally, we would like to extend our approach for unsupervised domain adaptation across multiple datasets.

\end{document}